# Processing Complex Sentences in the Centering Framework

**Michael Strube**

Freiburg University
Computational Linguistics Lab
Europaplatz 1, D-79085 Freiburg, Germany
strube@coling.uni-freiburg.de

## Abstract

We extend the centering model for the resolution of intra-sentential anaphora and specify how to handle complex sentences. An empirical evaluation indicates that the functional information structure guides the search for an antecedent within the sentence.

## 1 Introduction

The centering model (Grosz et al., 1995) focuses on the resolution of inter-sentential anaphora. Since intra-sentential anaphora occur at high rates in real-world texts, the model has to be extended for the resolution of anaphora at the sentence level. However, the centering framework is not fully specified to handle complex sentences (Suri & McCoy, 1994). This underspecification corresponds to the lack of a precise definition of the expression *utterance*, a term always used but intentionally left undefined[1]. Therefore, the centering algorithms currently under discussion are not able to handle naturally occurring discourse. Possible strategies for treating sentence-level anaphora within the centering framework are

1. processing sentences linearly one clause at a time (as suggested by Grosz et al. (1995)),
2. preference for sentence-external antecedents which are proposed by the centering mechanism,
3. preference for sentence-internal antecedents which are filtered by the usual binding criteria,
4. a mixed-mode which prefers only a particular set of sentence-internal over sentence-external antecedents (e.g. Suri & McCoy (1994)).

The question arises as to which strategy fits best for the interaction between the resolution of intra- and inter-sentential anaphora. In my contribution, evidence for a mixed-mode strategy is brought forward, which favors a particular set of sentence-internal antecedents given by functional criteria.

---
[1] Cf. the sketchy statements by Brennan et al. (1987, p.155): "[...] *U* is an utterance (not necessarily a full clause) [...]", and by Grosz et al. (1995, p.209): "*U* need not to be a full clause."

## 2 Constraints on Sentential Anaphora

Our studies on German texts have revealed that the functional information structure of the sentence, considered in terms of the context-boundedness of discourse elements, is the major determinant for the ranking on the *forward-looking-centers* ($C_f(U_n)$) (Strube & Hahn, 1996). Hence, context-bound discourse elements are generally ranked higher in the $C_f$ than any other non-anaphoric element. The functional information structure has impact not only on the resolution of inter-sentential anaphora, but also on the resolution of intra-sentential anaphora. Hence, the most preferred antecedent of an intra-sentential anaphor is a phrase which is also anaphoric. Consider sentences (1) and (2) and the corresponding centering data in Table 1 ($C_b$: backward-looking center; the first element of the pairs denotes the discourse entity, the second element the surface). In sentence (1), a nominal anaphor occurs, *der T3100SX* (a particular notebook). In sentence (2), another nominal anaphor appears, *der Rechner (the computer)*, which is resolved to T3100SX from the previous sentence. In the matrix clause, the pronoun *er (it)* co-specifies the already resolved anaphor *der Rechner* in the subordinate clause.

(1) Ist der Resume-Modus aktiviert, schaltet sich der *T3100SX* selbständig ab.
(If the resume mode is active, – switches – itself – the *T3100SX* – automatically – off.)

(2) Bei späterem Einschalten des *Rechners* arbeitet *er* sofort an der alten Stelle weiter.
(The – later – turning on – of the *computer* – it – resumes working – at exactly the same place.)

| (1) | **Cb:** | T3100SX: T3100SX |
|---|---|---|
|     | **Cf:** | [T3100SX: T3100SX] |
| (2) | **Cb:** | T3100SX: er |
|     | **Cf:** | [T3100SX: er, TURN-ON: Einschalten, PLACE: Stelle] |

Table 1: Centering Data for Sentences (1) and (2)

This example illustrates our hypothesis that intra-sentential anaphors preferably co-specify context-bound discourse elements. In order to empirically strenghten this argument, we have examined several texts of different types: 15 texts from the information technology (IT) domain, one text from the German news magazine *Der Spiegel*, and the first chapters of a short story by the German writer *Heiner Müller*[2] (cf. Table 2). In the texts, 65 intra-sentential anaphors oc-cur, 58 of them (89,2%) have an antecedent which is a resolved anaphor, while only 32 of them (49,2%) have an antecedent which is the subject of the matrix clause (cf. Table 3). These data indicate that an approach based on grammatical roles (Suri & McCoy, 1994) is inappropriate for the German language, while an approach based on the functional information structure seems preferable. In addition, we maintain that exchanging grammatical with functional criteria is also a reasonable strategy for fixed word order languages. They can be rephrased in terms of functional criteria, simply due to the fact that grammatical roles and the information structure patterns we defined, unless marked, coincide in these languages.

|         | text ana. | sent. ana. | anaphors | words |
|---------|-----------|------------|----------|-------|
| IT      | 284       | 24         | 308      | 5542  |
| Spiegel | 90        | 12         | 102      | 1468  |
| Müller  | 124       | 29         | 153      | 867   |
| $\Sigma$ | 498      | 65         | 563      | 7877  |

Table 2: Distribution of Anaphors in the Text Corpus

|         | cont.-bound | $\neg$ bound | subj. | $\neg$ subj. |
|---------|-------------|--------------|-------|--------------|
| IT      | 20          | 4            | 16    | 8            |
| Spiegel | 10          | 2            | 6     | 6            |
| Müller  | 28          | 1            | 10    | 19           |
| $\Sigma$ | 58         | 7            | 32    | 33           |

Table 3: Types of Intra-Sentential Antecedents

Since the strategy described above is valid only for complex sentences which consist of a matrix clause and one or more subordinate clauses, compound sentences which consist of main clauses must be considered. Each of these sentences is processed by our algorithm in linear order, one clause at a time with the usual centering operations. Compound sentences which consist of multiple full clauses also have multiple $C_b/C_f$ data.

Now, we are able to define the expression *utterance* in a satisfactory manner: An utterance $U$ is a simple sentence, a complex sentence, or each full clause of a compound sentence[3]. The $C_f$ of an utterance is computed only with respect to the matrix clause. Given these findings, complex sentences can be processed at three stages (2a-2c; transitions from one stage to the next occur only when a suitable antecedent has not been found at the previous stage):

---

[2]Liebesgeschichte. In Heiner Müller, *Geschichten aus der Produktion 2*, Berlin: Rotbuch Verlag, pp.57-63.

[3]We do not consider dialogues with elliptical utterances.

1. For resolving an anaphor in the first clause of $U_n$, propose the elements of $C_f(U_{n-1})$ in the given order.

2. For resolving an anaphor in a subsequent clause of $U_n$,
    (a) propose already context-bound elements of $U_n$ from left to right[4].
    (b) propose the elements of $C_f(U_{n-1})$ in the given order.
    (c) propose all elements of $U_n$ not yet checked from left to right.

3. Compute the $C_f(U_n)$, considering only the elements of the matrix clause of $U_n$.

## 3  Evaluation

In order to evaluate the functional approach to the resolution of intra-sentential anaphora within the centering model, we compared it to the other approaches mentioned in Section 1, employing the test set referred to in Table 2. Note that we tried to eliminate error chaining and false positives (for some remarks on evaluating discourse processing algorithms, cf. Walker (1989); we consider her results as a starting point for our proposal).

First, we examine the errors which all strategies have in common (for the success rate, cf. Table 4). 99 errors are caused by underspecification at different levels, e.g., prepositional anaphors (16), plural anaphors (8), anaphors which refer to a member of a set (14), sentence anaphors (21), and anaphors which refer to a global focus (12) are not yet included in the mechanism. In 9 cases, any strategy will choose the false antecedent.

The most interesting cases are the ones for which the performance of the different strategies varies. The linear approach generates 40 additional errors in the anaphora resolution, which are caused only by the ordering strategy to process each clause of sentences with the centering mechanism. The approach which prefers inter-sentential anaphora causes 60 additional errors. Note that this strategy performs remarkably well at first sight. For 44 of the errors it chooses an inter-sentential antecedent which is, on the surface, identical to the correct intra-sentential antecedent. We count these 44 resolutions as false positives, since the anaphor has been resolved to the false discourse entity. The approach which prefers intra-sentential antecedents causes 27 additional errors. These errors occur whenever an inter-sentential anaphor can be resolved with an incorrect intra-sentential antecedent.

---

[4]We abstract here from the syntactic criteria for filtering out some elements of the current sentence by applying binding criteria (Strube & Hahn, 1995). Syntactic constraints like control phenomena override the preferences given by the context.

|         | N   | linear |         | extra > intra |         | intra > extra |         | functional |         |
|---------|-----|--------|---------|---------------|---------|---------------|---------|------------|---------|
| IT      | 308 | 237    | (76,9%) | 229           | (74,3%) | 240           | (77,9%) | 247        | (80,2%) |
| Spiegel | 102 | 82     | (80,4%) | 76            | (74,5%) | 82            | (80,4%) | 86         | (84,3%) |
| Müller  | 153 | 105    | (68,8%) | 99            | (64,7%) | 115           | (75,3%) | 128        | (83,7%) |
| Σ       | 563 | 424    | (75,3%) | 404           | (71,8%) | 437           | (77,6%) | 461        | (81,9%) |

Table 4: Success Rate without Semantic Constraints

The functional approach causes only 3 additional errors. These errors occur whenever the antecedent of an intra-sentential anaphor is not bound by the context (which is possible but rare) and when the anaphor can be resolved at the text level.

The results change slightly if semantic/conceptual constraints (type and further admissibility constraints) on anaphora are considered. 22 errors of the linear approach, 8 errors of the approach which prefers inter-sentential antecedents, and 12 errors of the approach which prefers inter-sentential antecedents can be avoided. Only 6 errors of the functional approach can be avoided by incorporating semantic criteria. This might constitute a cognitively valid argument for the functional approach – the better the strategy, the lower the influence of semantics or world knowledge on anaphora resolution.

To summarize the results of our empirical evaluation, we claim that our proposal based on functional criteria leads to substantively better results for languages with free word order than the linear approach suggested by Grosz et al. (1995) and the two approaches which prefer inter-sentential or intra-sentential antecedents.

## 4 Comparison to Related Work

Crucial for the evaluation of the centering model (Grosz et al., 1995) and its applicability to naturally occurring discourse is the lack of a specification concerning how to handle complex sentences and intra-sentential anaphora. Grosz et al. suggest the processing of sentences linearly one clause at a time. We have shown that such an approach is not appropriate for some types of complex sentences. Suri & McCoy (1994) argue in the same manner, but we consider the functional approach for languages with free word order superior to their grammatical criteria, while, for languages with fixed word order, both approaches should give the same results. Hence, our approach seems to be more generally applicable. Other approaches which integrate the resolution of sentence- and text-level anaphora are based on salience metrics (Hajičová et al., 1992; Lappin & Leass, 1994). We consider such metrics to be a method which detracts from the exact linguistic specifications as we propose them.

At first sight, grammar theories like GB (Chomsky, 1981) or HPSG (Pollard & Sag, 1994), are the best choice for resolving anaphora at the sentence-level. But these grammar theories only give filters for excluding some elements from consideration. Neither gives any preference for a particular antecedent at the sentence-level, nor do they consider text anaphora.

## 5 Conclusions

In this paper, we gave a specification for handling complex sentences in the centering model based on the functional information structure of utterances in discourse. We motivated our proposal by the constraints which hold for a free word order language (German) and derived our results from data-intensive empirical studies of real texts of different types.

Some issues remain open: the evaluation of the functional approach for languages with fixed word order, a fine-grained analysis of subordinate clauses as Suri & McCoy (1994) presented for *SX because SY* clauses, and, in general, the solution for the cases which cause errors in our evaluation.

**Acknowledgments.** This work has been funded by *LGFG Baden-Württemberg*. I would like to thank my colleagues in the $\mathcal{CLIF}$ group for fruitful discussions. I would also like to thank Jon Alcantara (Cambridge) who kindly took the role of the native speaker via Internet.

## References

Brennan, S. E., M. W. Friedman & C. J. Pollard (1987). A centering approach to pronouns. In *Proc. of ACL-87*, pp. 155–162.

Chomsky, N. (1981). *Lectures on Government and Binding*. Dordrecht: Foris.

Grosz, B. J., A. K. Joshi & S. Weinstein (1995). Centering: A framework for modeling the local coherence of discourse. *Computational Linguistics*, 21(2):203–225.

Hajičová, E., V. Kuboň & P. Kuboň (1992). Stock of shared knowledge: A tool for solving pronominal anaphora. In *Proc. of COLING-92*, Vol. 1, pp. 127–133.

Lappin, S. & H. J. Leass (1994). An algorithm for pronominal anaphora resolution. *Computational Linguistics*, 20(4):535–561.

Pollard, C. & I. V. Sag (1994). *Head-Driven Phrase Structure Grammar*. Chicago, ILL: Chicago Univ. Press.

Strube, M. & U. Hahn (1995). *ParseTalk* about sentence- and text-level anaphora. In *Proc. of EACL-95*, pp. 237–244.

Strube, M. & U. Hahn (1996). Functional centering. In *Proc. of ACL-96*.

Suri, L. Z. & K. F. McCoy (1994). RAFT/RAPR and centering: A comparison and discussion of problems related to processing complex sentences. *Computational Linguistics*, 20(2):301–317.

Walker, M. A. (1989). Evaluating discourse processing algorithms. In *Proc. of ACL-89*, pp. 251–261.